\documentclass[11pt]{article}

\usepackage{amsmath}
\usepackage{hyperref}
\usepackage{algorithm}      
\usepackage{algorithmic}
\usepackage{graphicx}
\usepackage{booktabs}
\usepackage{multicol}
\usepackage{tabularx}
\usepackage{enumitem}
\usepackage{multirow}
\usepackage{xcolor}
\usepackage{subcaption}
\usepackage{amsmath}
\usepackage{amsfonts}
\usepackage{enumitem}
\usepackage{stfloats}
\usepackage[preprint]{acl}

\usepackage{times}
\usepackage{latexsym}

\usepackage[T1]{fontenc}

\usepackage[utf8]{inputenc}

\usepackage{microtype}

\usepackage{inconsolata}

\usepackage{graphicx}

\title{KV Pareto: Systems-Level Optimization of KV Cache and Model Compression for Long Context Inference}

\author{
Sai Gokhale \thanks{This work was done during her internship with Advanced Micro Devices (AMD)} \thanks{Equal Contribution} \\ Georgia Institute of Technology
\AND
Devleena Das \footnotemark[2] \quad Rajeev Patwari \footnotemark[2] \quad Ashish Sirasao \quad Elliott Delaye \\ Advanced Micro Devices (AMD)
}

\begin{document}
\maketitle
\begin{abstract}
Long-context Large Language Models (LLMs) face significant memory bottlenecks during inference due to the linear growth of key-value (KV) cache with sequence length. While individual optimization techniques like KV cache quantization, chunked prefill, and model weight quantization have shown promise, their joint effects and optimal configurations for edge deployment remain underexplored. We introduce KV Pareto, a systems-level framework that systematically maps the trade-off frontier between total memory consumption and task accuracy across these three complementary optimization techniques. Our framework evaluates multiple LLM architectures (Qwen, Llama, Mistral) with varying KV quantization schemes (int2/4/8, mixed-precision), granularities (per-token, per-tensor, per-block), and 4-bit weight quantization via AWQ. Our framework identifies model-specific Pareto-optimal configurations that achieve 68-78\% total memory reduction with minimal (1-3\%) accuracy degradation on long-context tasks. We additionally verify the selected frontiers on additional benchmarks of Needle-in-a-Haystack, GSM8k and MMLU as well as extended context lengths of up to 128k to demonstrate the practical need of joint optimization for efficient LLM inference.
\end{abstract}

\section{Introduction}
\label{introduction}
Large Language Models (LLMs) have become useful in many applications, such as code generation \cite{jiang2024survey}, long question-answering \cite{liu2025comprehensive} and retrieval-augmented generation (RAG) \cite{arslan2024survey}. 
These tasks increasingly demand longer-context capabilities, pushing models like Qwen \cite{bai2023qwen}, Mistral \cite{jiang2024mistral} and Llama \cite{grattafiori2024llama} to support long context lengths.


The bottleneck for efficient inference arises from the transformer architecture which operates primarily in two phases: prefill and decode \cite{raiaan2024review}. During prefill, the input context is processed and stored in the key-value (KV) cache. During decode, outputs are generated autoregressively by repeatedly accessing the KV cache. Importantly, the KV Cache size grows linearly with sequence length \cite{patwari2025forecasting} and increases time-to-first-token (TTFT) during prefill, and time-per-output-token (TPOT) during decode. The resulting increased latency is a bottleneck for practical deployment of long-context models on edge devices.

To reduce the latency and scalability challenges, several optimization techniques have been proposed for long-context inference, including KV Quantization \cite{hooper2024kvquant,li2025kvtuner,liu2024kivi}, token eviction \cite{xiao2023efficient, corallo2024finch}, and chunked prefill \cite{agrawal2023sarathiefficientllminference}. These methods are typically evaluated in isolation of one another in the context of accuracy degradation \cite{li2025kvtuner,liu2024kivi, corallo2024finch}. This creates a practical question for deployment: \textit{which memory optimizations, together, offer the best-trade offs between memory savings and task accuracy?} 

To this end, we introduce \textbf{KV Pareto}, a framework for evaluating and understanding the trade-offs between KV memory compression and task performance in long-context LLMs. KV Pareto focuses on studying the impact of two widely accessible optimization techniques for long context, KV quantization, and chunked prefill in conjunction with 4-bit model weight quantization. Prior work evaluates these optimization techniques in isolation \cite{li2025kvtuner,liu2024kivi, corallo2024finch, lin2024awq,agrawal2023sarathiefficientllminference}. Instead, our KV Pareto provides a joint assessment of optimization techniques, considering total memory savings and accuracy degradation. This enables practitioners to identify the Pareto-optimal configurations for edge deployment.

Our KV Pareto spans multiple models (Mistral, Qwen, LLaMA), KV cache quantization granularities (per-token, per-tensor, per-block), group sizes (32, 64, 128),  precision formats (int2, int4, int8), as well 4-bit weight quantization via AWQ \cite{lin2024awq}. We benchmark across long context evaluations including LongBench \cite{bai2024longbench}, Needle-in-a-Haystack (NIAH) \cite{NIAH} , and traditional tasks such as GSM8k \cite{gsm8k} and MMLU \cite{mmlu}, measuring total memory through peak activation memory, KV memory, and model memory, as well as task accuracy. Our contributions are:
\begin{enumerate}
    \item \textbf{KV Pareto Framework:} A KV optimization Pareto framework that systematically maps the trade-off search space between total memory savings and task accuracy.
    \item \textbf{Joint Optimization Study:} A comprehensive evaluation of chunked prefill, KV cache quantization and AWQ weight-only quantization across multiple KV cache quantization granularities and precisions, identifying Pareto-optimal configurations using LongBench.
    \item \textbf{KV Pareto Validation:} Validation our framework's selected frontiers on NIAH, showing strong task performance even at 20-32k context lengths, as well as MMLU, GSM8k.
\end{enumerate}

\section{Related Works}
\label{related works}

\subsection{KV Quantization}
KV quantization reduces the precision of stored key and value tensors, thereby lowering memory usage \cite{li2024survey}. For example, KIVI \cite{liu2024kivi} and KVQuant \cite{hooper2024kvquant} introduce tuning-free asymmetric quantization schemes that apply per-channel and per-token quantization, achieving up to 2-bit compression. 
More recently, KVTuner \cite{li2025kvtuner} proposes an adaptive framework that searches for the optimal layer-wise KV quantization precision pairs and demonstrates near lossless 3.25 bit mixed precision KV quantization for mathematical reasoning tasks. Inspired by KVTuner \cite{li2025kvtuner}, our KV Pareto framework also considers mixed-precision quantization schemes. While KVTuner \cite{li2025kvtuner} focuses on layerwise adaptability, our KV Pareto framework focuses on additional important quantization hyperparameters such as, quantization scheme (blockwise, per tensor, per token), as well as system-level interactions with other optimizations such as, PC and model quantization.

\subsection{Prefill Chunking}
Prefill chunking (PC) reduces the peak memory consumption by dividing input prompts into equal-sized segments that are processed sequentially \cite{agrawal2023sarathiefficientllminference}. PC is adopted in inference systems like vLLM \cite{kwon2023efficient}. Additionally, follow on works such as WiM \cite{russak2024writing} leverage the concept of smaller chunks to improve model reasoning.
However, the benefits and effects of PC has not yet been studied in conjunction with model quantization as well as KV cache quantization. We address this gap by analyzing PC at a systems-level, understanding its tradeoffs when combined with KV cache and model quantization.


\subsection{Model Quantization}
Model quantization algorithms are popular for efficient inference as they significantly compress model size. 
Popular methods include, GPTQ \cite{frantar2023gptqaccurateposttrainingquantization}, and AWQ \cite{lin2024awq}. GPTQ \cite{frantar2023gptqaccurateposttrainingquantization} uses second-order information derived from the Hessian matrix to enable 3-4 bit quantization. AWQ \cite{lin2024awq} preserves accuracy in 4-bit quantization by using activation metrics to identify the most important weight channels, which are then scaled prior to quantization to reduce error. In our framework, we study model memory savings via AWQ \cite{lin2024awq} to provide practical insights for edge deployment on how model compression, with KV quantization and PC, can provide the most memory savings with the least task performance degradation.

\section{Background}
\label{background}
The KV cache is a crucial component for LLM inference, storing intermediate representations that are used for autoregressive generation. KV cache memory can be represented as the follows: 
$\text{KV Cache Memory} = B \times H \times N \times D \times L \times s$, where $B$ is batch size, $H$ is number of attention heads, $N$ is number of tokens stored in the cache, $L$ is number of layers, $D$ is head dimension and $s$ is the size per element.


At a systems level, KV cache optimizations for edge deployment remain largely unexplored when considering its interactions with other memory-savings optimizations such as weight-only quantization and prefill chunking. Therefore, our work focuses specifically on the \textit{joint} interactions among KV cache quantization, prefill chunking, and model weight quantization, on accuracy and total memory.




\subsection{KV Quantization Optimization}
KV Quantization reduces the element size $s$ by using lower-precision formats (e.g. int8, int4, int2), allowing memory savings: 
$M^{quant}_{KV} << M^{bf16}_{KV}$. Additionally, quantizing the KV cache reduces memory bandwidth, improving TPOT during decode. However, KV cache quantization also introduces approximation error $\epsilon_{Q_{KV}}$, which can degrade attention quality and therefore task accuracy. 

\subsection{Prefill Chunking Optimization}
\label{sec:PC-background}
Standard prefill involves processing the entire input $M$ in one pass, which leads to higher peak memory consumption due to the size of the attention computation. The attention weights are computed as: $softmax(\left(\frac{QK^\top}{\sqrt{d_k}}\right))$, where $Q$, $K$ are the query and key matrices, and $d_{k}$ is dimensionality of the key matrix \cite{attention_is_all_you_need}. The larger the $M$, the larger the query matrix $Q$, and therefore larger the attention computation and associated peak memory. Thus, peak memory during prefill can loosely be approximated as: $M_{peak} \approx M_{atten}$.

PC reduces the peak memory by dividing $M$ into smaller chunks sizes of $k$ << $M$, and processing each chunk sequentially. This limits the number of queries computed at once, thereby reducing the size of the attention computation: $M^{chunked}_{peak} \approx \max_i (M_{atten}(k_{i}))$, where $k_{i}$ represents the number of tokens in chunk $i$. 

\subsection{Model Weight Optimization}
While KV cache optimization is crucial, model weight quantization is often needed for deploying larger models on edge devices with constrained memory. Therefore, we also consider AWQ \cite{lin2024awq}, a SoTA weight-only quantization technique which further reduces the total memory. However, the KV cache quantization error $\epsilon_{Q_{KV}}$ and AWQ weight quantization $\epsilon_{Q_{W}}$ can compound errors, further degrading task performance. 

Each type of optimization presents its own set of hyperparameters, and optimizing across these require a systematic framework. Our KV Pareto empirically identifies the Pareto frontier for a given model, characterizing tradeoffs between total memory and accuracy for long-context inference.

\begin{figure*}[ht]
    \centering
    \captionsetup{justification=centering}
    \includegraphics[width=1.0\textwidth]{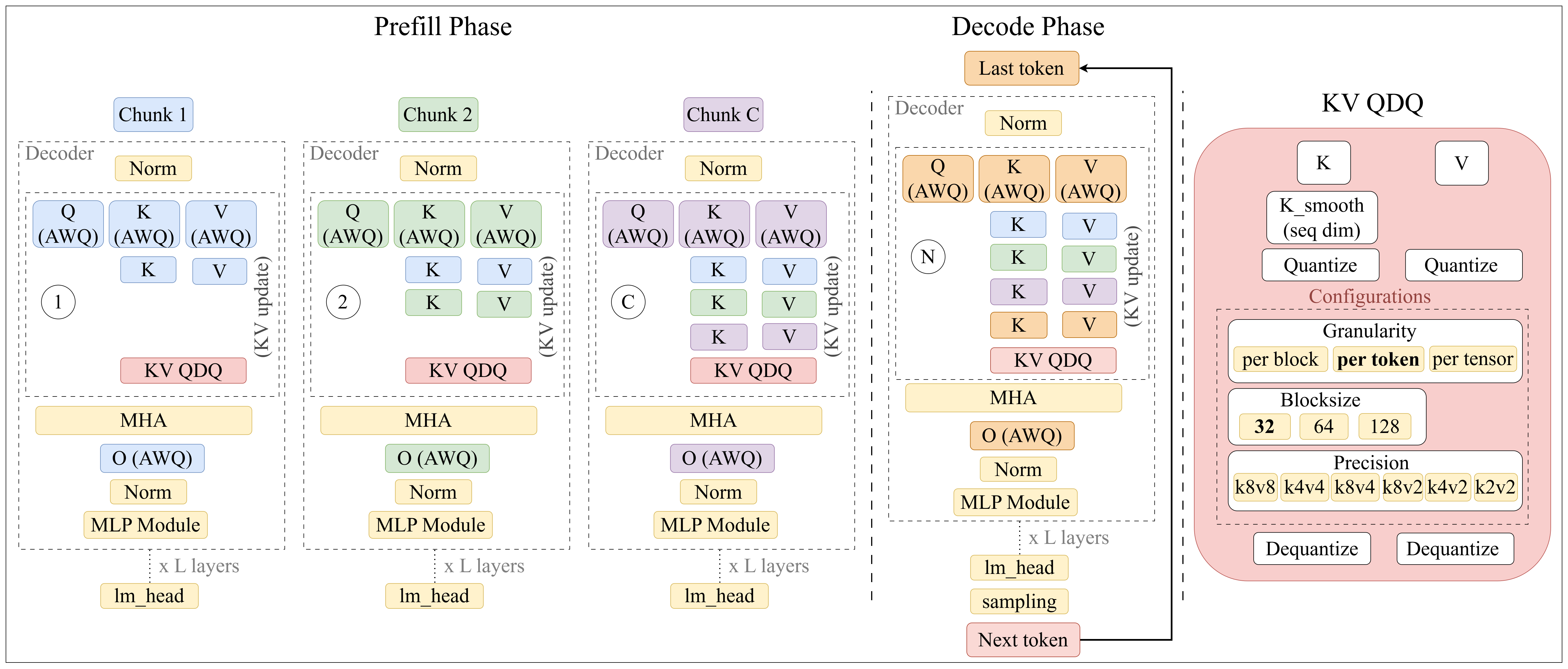}
    \caption{Our KV Pareto Framework, showcasing the integration of prefill chunking (PC), KV cache quantization and model quantization for prefill and decode phases.}
    \label{fig:kvpareto_method}
\end{figure*}

\section{Methodology}
\label{methodology}
Our KV Pareto Framework  
finds the Pareto frontiers, per model, considering the trade-offs between total memory savings and task accuracy. KV Pareto is designed to find the frontier, from a systems level, considering not only KV cache quantization schemes, but additionally further memory savings from PC and model-weight quantization. As shown in Figure \ref{fig:kvpareto_method}, PC is enabled in the prefill phase where a prompt of length $M$ is segmented into $C$ smaller chunks, such that the size of the Query matrix $Q$ into the multi-head-attention block (MHA) is of size $c_{i} \in C$, lowering peak memory consumption. KV Cache quantization is enabled both in the prefill and decode phase, and simulated by a quantization-dequantization (QDQ) process to insert quantization error into both the key $K$ and value $V$ states. Lastly, weight quantization can be applied via AWQ for both the prefill and decode phase on all linear layers.






\subsection{Prefill Chunking}
PC partitions the input into equal-sized chunks, and the KV cache is filled iteratively. As shown in Figure \ref{fig:kvpareto_method}, when the prompt is divided into \(C\) chunks, the KV cache undergoes \(C\) updates. For each chunk \(i \in \{1, \dots, C\}\), the update can be expressed as: $\text{KV}_{i} \leftarrow \text{KV}_{i-1} + \{K_i, V_i\}$. As mentioned in Section \ref{sec:PC-background}, PC lowers the size of the attention computation, thereby reducing peak memory consumption. We ran ablations to understand the impact of different chunk sizes, $\{64, 128, 256, 512, 1024\}$, on task accuracy. Appendix \ref{sec:PC-ablations} shows minimal accuracy changes from PC, and for consistent comparisons within our framework, we set chunk size to 256 for all KV Pareto experiments.


\subsection{KV Quantization}
KV quantization compresses the K and V matrices into a lower bit width. We consider int8, int4 and int2 quantization with mixed-precision variants: $\{k8v8, k8v4, k8v2, k4v4, k4v2, k2v2\}$ and apply signed, asymmetric round-to-nearest (RTN) quantization. Appendix \ref{sec:RTN-eqs} provides details on RTN quantization using 3 techniques,\textbf{ per-token groupwise, per-sequence groupwise and per-tensor}. We also apply k-smoothing \cite{zhang2025sageattention} to improve quantization error. Appendix \ref{sec:kv-quant-ablations} shows our ablations on varying group sizes, showing larger models perform best at per-token, group size 64, and smaller models at, per token, group size 32. 

\paragraph{K-smoothing} Inspired by SageAttention \cite{zhang2025sageattention}, we apply mean smoothing to the $K$ tensor, prior to quantization, to mitigate uneven distributions in $K$. Appendix \ref{sec:k-smoothing-method} details the K-smoothing process and our ablations reveal k4v4 significantly benefits from K-smoothing.

\subsection{Model Quantization}
KV Pareto also considers the benefit of model weight compression via weight-only quantization to reduce total memory utilization. Given its SoTA performance, we apply AWQ \cite{lin2024awq}, which selectively protects important weight channels based on activation statistics calculated from calibration data. We leverage a robust configuration of AWQ: 4-bit unsigned, asymmetric quantization with group size 128 along the channel dimension.

\section{Experimental Design}
All experiments are performed on AMD MI-210 and MI-325 GPUs. 

\textbf{Datasets.} We evaluate long context performance with Hotpotqa \cite{yang2018hotpotqadatasetdiverseexplainable} and Qasper \cite{qasper} from LongBench \cite{bai2024longbenchbilingualmultitaskbenchmark}. To ensure KV Pareto does not degrade shorter-context tasks, we also evaluate on GSM8k \cite{gsm8k} and MMLU \cite{mmlu}. Dataset details are in Appendix \ref{sec:eval-datasets}.

\textbf{Models.} We evaluate across diverse LLM architectures, including Qwen2.5-3b and Qwen2.5-7b instruct \cite{qwen2025qwen25technicalreport}, Llama3.2-3b and Llama3.1-8b instruct \cite{grattafiori2024llama3herdmodels}, and Mistral-7b-instruct-v0.3 \cite{jiang2023mistral7b}. 

\subsection{KV Pareto Frontier Metrics}
In our context, a configuration is Pareto dominated if there exists another configuration that achieves equal or better task performance with lesser total memory utilization. Our metrics are:
\begin{enumerate}
    \item \textbf{Total Memory Consumption} We approximate total memory utilization to include \textit{peak memory, KV cache memory, and model memory}. Appendix \ref{sec:mem-util-calc} provides details.
    \item \textbf{Task Accuracy} We measure how accurate each Pareto configuration is on the LongBench \cite{bai2024longbench} tasks to analyze impact of joint optimizations on task performance.
\end{enumerate}

\textbf{KV Pareto Validation.} We validate our selected pareto-optimal configurations using NIAH \cite{NIAH}, GSM8k\cite{gsm8k} and MMLU\cite{mmlu} to ensure robustness at higher context lengths and non-long context tasks. \cite{yang2018hotpotqadatasetdiverseexplainable}. 




\begin{figure*}[t!]
    \centering
    \captionsetup{justification=centering}
    \begin{subfigure}[b]{0.48\textwidth}
        \includegraphics[width=1.0\linewidth]{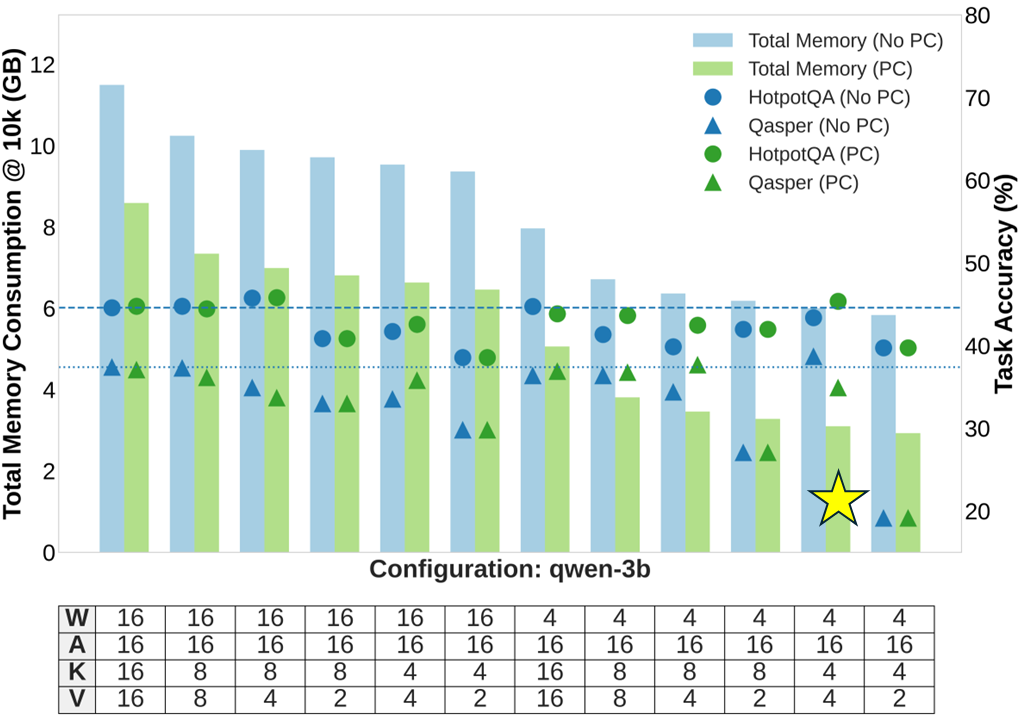}
        \caption{Qwen-2.5-3B-I's pareto frontier is w4a16\_k4v4}
        \label{fig:pareto_slide7}
    \end{subfigure}
    \hfill
    \begin{subfigure}[b]{0.48\textwidth}
        \includegraphics[width=1.0\linewidth]{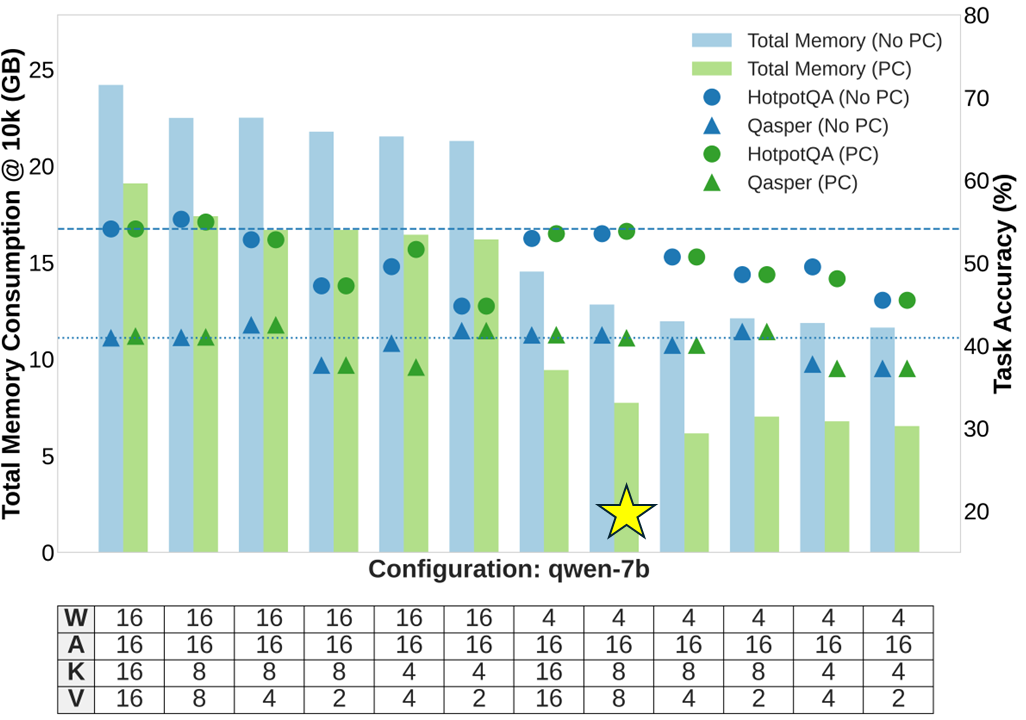}
        \caption{Qwen-2.5-7B-I's pareto frontier is w4a16\_k8v8}
        \label{fig:pareto_slide3}
    \end{subfigure}

    \begin{subfigure}[b]{0.48\textwidth}
        \includegraphics[width=\linewidth]{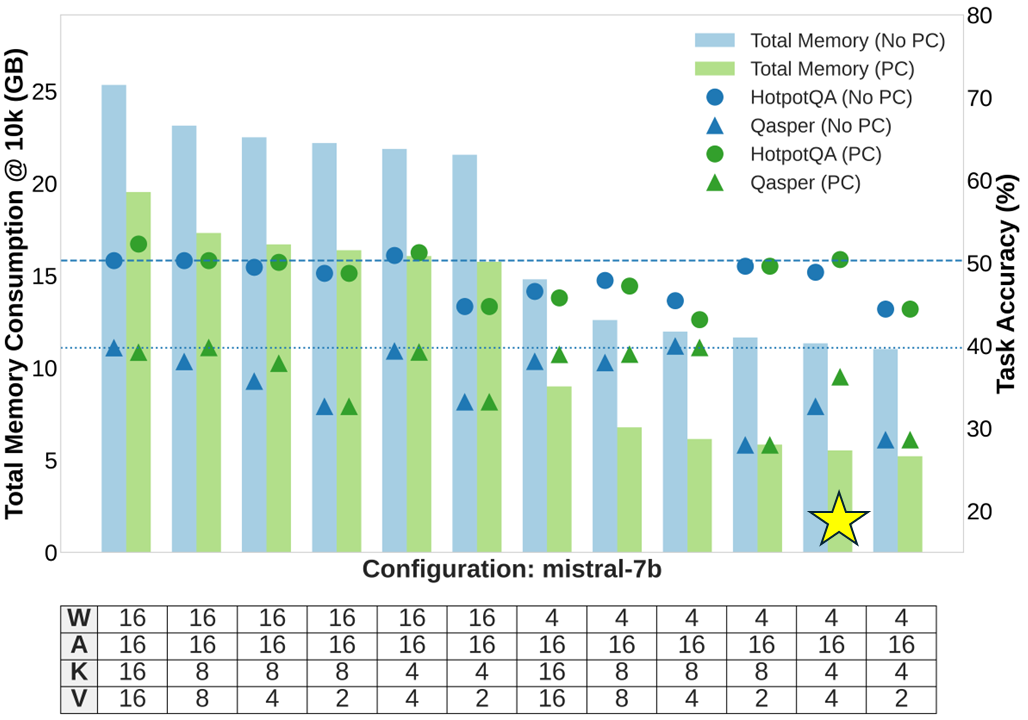}
        \caption{Mistral-v0.3-7B-I's pareto frontier is w4a16\_k4v4}
        \label{fig:pareto_slide6}
    \end{subfigure}
    \hfill
    \begin{subfigure}[b]{0.48\textwidth}
        \includegraphics[width=1.0\linewidth]{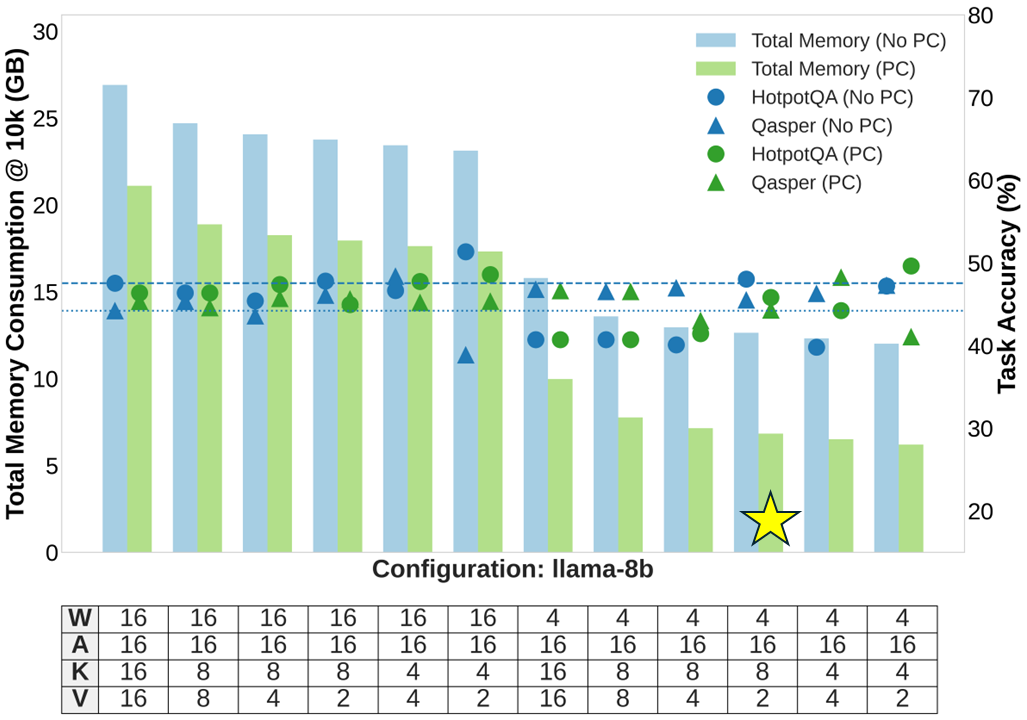}
        \caption{Llama-3.2-8B-I's pareto frontier is w4a16\_k8v2}
        \label{fig:pareto_slide5}
    \end{subfigure}


    \caption{Pareto curves for five models that show the tradeoff between task accuracy and memory consumption, with frontiers shown with a star, and horizontal lines showing baseline (w16a16\_k16v16) accuracy. }
    \label{fig:pareto_grid}
\end{figure*}

\section{Results}
\subsection{KV Pareto Frontiers}
Figure \ref{fig:pareto_grid} shows the Pareto-optimal configurations from our framework, measuring total memory consumption at a 10k context length alongside accuracies of two long-context tasks. The pareto frontiers yield 68-78\% memory savings with marginal (1-3\%) task accuracy drop. The frontier for Qwen2.5-3b-instruct, Mistral-v0.3-7b-instruct and Llama-3.2-3b-instruct (See Appendix \ref{sec:add-pareto-curves}) is w4a16-k4v4,  while for Qwen2.5-7b-instruct it is w4a16-k8v8, and for Llama3.2-8b-instruct it is w4a16-k8v2. 

\paragraph{Benefit of Joint Study} From Figure \ref{fig:pareto_grid}, we see that PC yields the most reduction in peak memory with minimal changes to task accuracy and AWQ further reduces memory consumption. While AWQ generally causes task accuracy loss, there are instances where it benefits task accuracy. For example, pairing 4-bit weight quantization with k4v4 improves HotpotQA accuracy compared to k8v4. Similarly, combining PC with KV quantization yields higher-than-baseline task accuracies on Qasper, while reducing memory footprint (w16a16-k16v16 vs w16a16-k8v8). We hypothesize this improvement stems from k-smoothing. Our findings stress the importance of our framework, and considering Pareto-optimal configurations at a systems-level for edge deployment to maximize tradeoffs.

\begin{table*}[t!]
\centering
\captionsetup{justification=centering}
\large 
\setlength{\tabcolsep}{3pt} 
\renewcommand{\arraystretch}{1.0} 
\resizebox{\textwidth}{!}{
\begin{tabular}{@{} c c | ccc | ccc | ccc | ccc | ccc @{}}
\toprule
\textbf{PC} & \textbf{AWQ} & \multicolumn{3}{c|}{\textbf{Qwen2.5-3B-I}} & \multicolumn{3}{c|}{\textbf{Qwen2.5-7B-I}} & \multicolumn{3}{c|}{\textbf{Llama-3.2-3B-I}} & \multicolumn{3}{c|}{\textbf{Llama-3.1-8B-I}} & \multicolumn{3}{c@{}}{\textbf{Mistral-v0.3-7B-I}} \\
\cmidrule(lr){3-5} \cmidrule(lr){6-8} \cmidrule(lr){9-11} \cmidrule(lr){12-14} \cmidrule(lr){15-17}
 & & K/V & gsm8k & mmlu & K/V & gsm8k & mmlu & K/V & gsm8k & mmlu & K/V & gsm8k & mmlu & K/V & gsm8k & mmlu \\
\midrule
no  & no  & 16/16 & 60.95 & 66.91 & 16/16 & 77.48 & 70.00 & 16/16 & 68.76 & 58.45 & 16/16 & 78.92 & 64.52 & 16/16 & 50.79 & 60.94 \\
yes & no  & 16/16 & 61.48 & 66.87 & 16/16 & 77.17 & 70.07 & 16/16 & 68.84 & 58.66 & 16/16 & 76.50 & 64.50 & 16/16 & 50.03 & 60.98 \\
yes & no & 4/4 & 56.63 & 65.95 & 8/8 & 77.28 & 69.96 & 4/4 & 67.55 & 57.33 & 8/2 & 66.00 & 60.40 & 4/4 & 50.64 & 60.17 \\
yes & yes & 16/16 & 60.12 & 61.92 & 16/16 & 71.03 & 69.64 & 16/16 & 51.78 & 56.07 & 16/16 & 75.74 & 64.30 & 16/16 & 48.30 & 60.49 \\
yes & yes & \textbf{4/4} & 59.21 & 61.33 & \textbf{8/8} & 71.72 & 69.64 & \textbf{4/4} & 61.03 & 57.51 & \textbf{8/2} & 66.00 & 60.28 & \textbf{4/4} & 43.66 & 58.77 \\
\bottomrule
\end{tabular}
}
\caption{Performance comparison PC, AWQ and selected pareto optimal configurations (bolded).}
\label{tab:pareto_gsm8k_mmlu}
\end{table*}

\subsection{Validation of KV Pareto Frontiers}
We validate the efficacy of our selected frontiers by further evaluating them on the following: 

\begin{figure*}[t!]
    \centering
    \captionsetup{justification=centering}
    \begin{subfigure}[b]{0.49\textwidth}
        \includegraphics[width=1.0\linewidth]{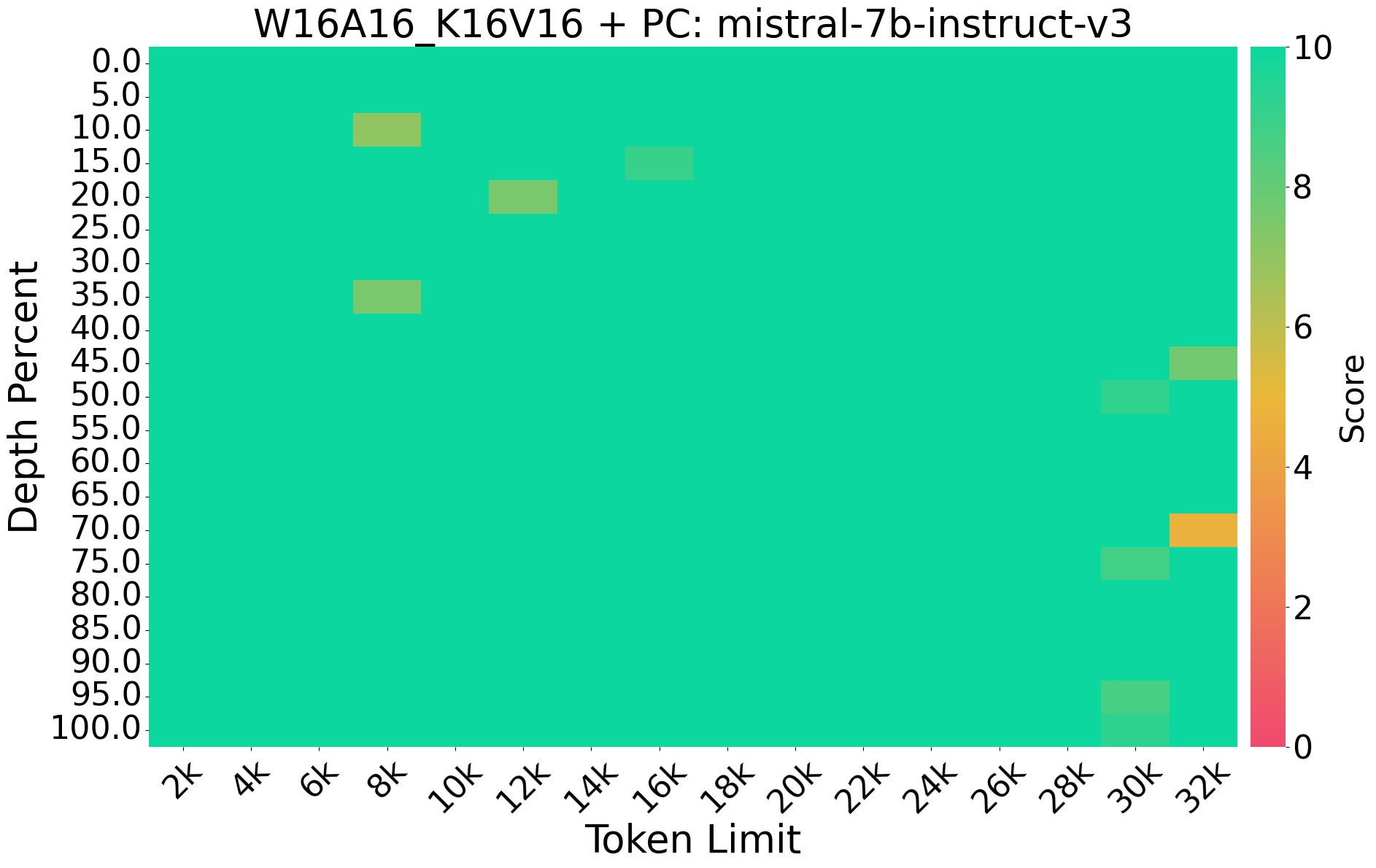}
        \caption{W16A16\_K16V16 retrieval scores for Mistral-v0.3-7B-I}
        \label{fig:niah1}
    \end{subfigure}
    \hfill
    \begin{subfigure}[b]{0.49\textwidth}
        \includegraphics[width=1.0\linewidth]{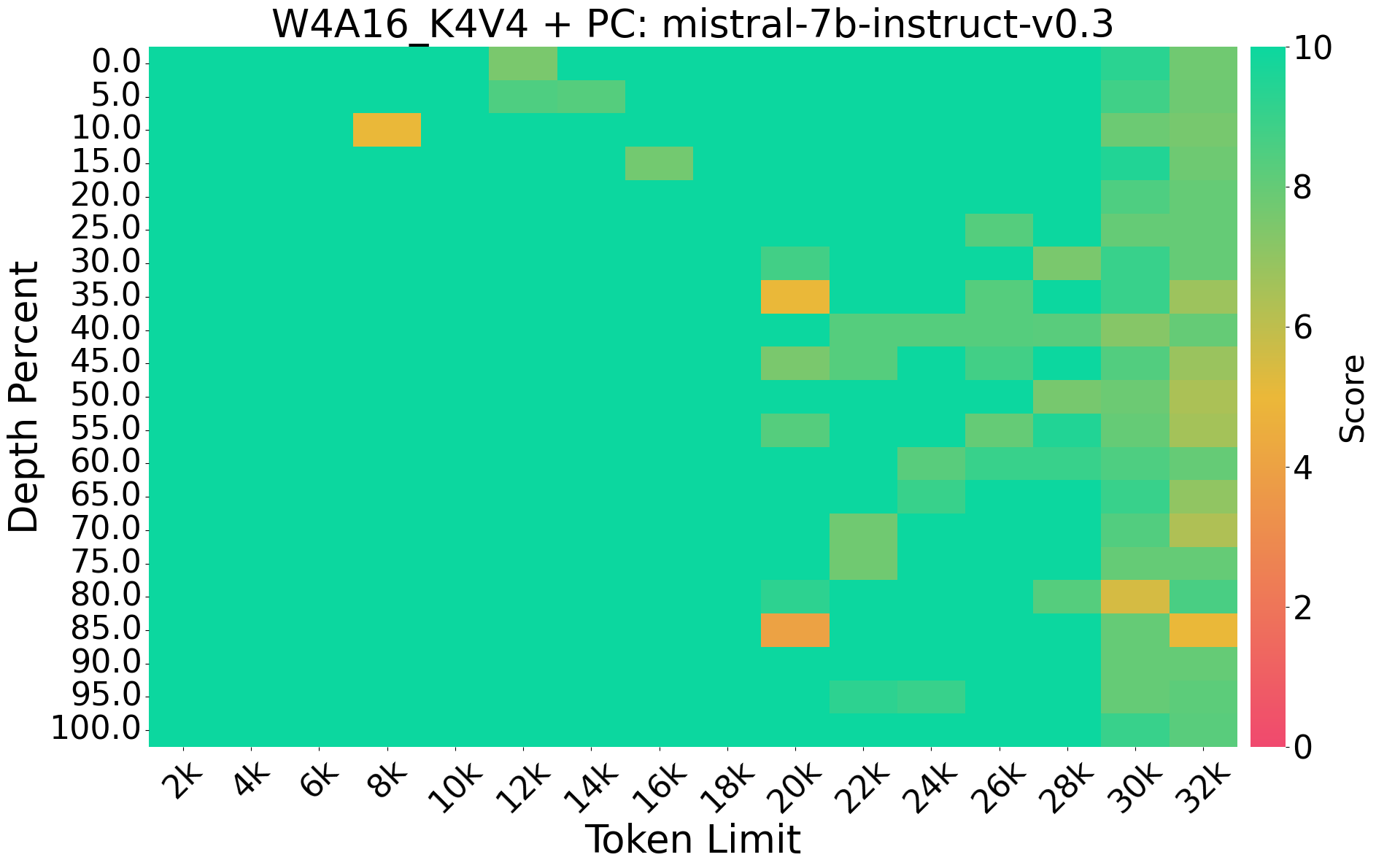}
        \caption{W4A16\_K4V4 retreival scores for Mistral-v0.3-7B-I} 
        \label{fig:niah2}
    \end{subfigure}


    \caption{NIAH performance on baseline (a) and pareto-optimal configurations (b).}
    \label{fig:niah_grid}
\end{figure*}

\paragraph{GSM8k \& MMLU Evaluations} Table \ref{tab:pareto_gsm8k_mmlu} shows the task accuracy for each pareto frontier on GSM8k and MMLU. Overall, GSM8k shows a greater performance drop compared to MMLU, with AWQ weight quantization having a stronger impact on GSM8k. In general, these results confirm the efficacy of our pareto-optimal configurations for complex, shorter context generation (GSM8k), and standard non-generation (MMLU) tasks, with 1-10\% degradation, depending on the model.

\paragraph{NIAH Evaluations} Figure \ref{fig:niah_grid} shows the retrieval scores for each depth (y axis) within a given document length (x-axis) for Mistral-v0.3-7b. The w4a16-k4v4 frontier maintains stable performance up to 20k tokens. These results suggest that beyond 20k, additional finetuning may be required to recover task accuracy while preserving memory savings. See Appendix \ref{sec:niah-results} for more NIAH results.

\subsection{Memory Savings Benefit Beyond 30k}
\begin{figure}[t]
    \centering
    \captionsetup{justification=centering}
    \includegraphics[width=1\linewidth]{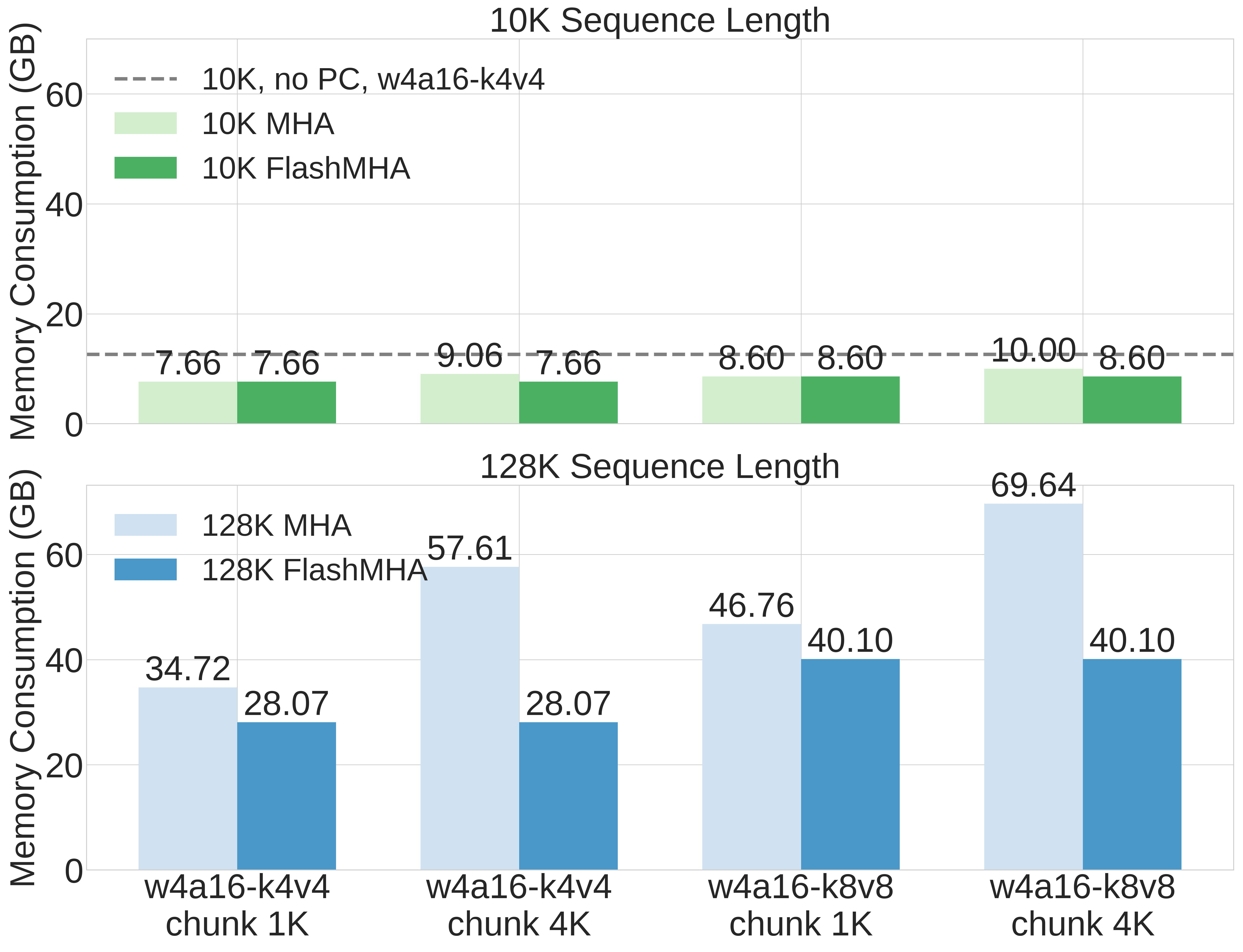}
    \caption{Peak memory consumption on 10k vs. 128k context lengths, comparing SDPA and Flash MHA.}
    \label{fig:mem_consumption_128k}
\end{figure}

Many real world applications, such as coding \cite{jiang2024survey} and RAG \cite{arslan2024survey}, require even larger context lengths. To address these practical scenarios, we analyze the benefit of our selected frontiers at extended context lengths, such as 128k tokens. Figure \ref{fig:mem_consumption_128k} explains the importance of taking a systems-level approach for selecting the pareto frontier, as each additional optimization provides a significant memory savings. For example, a smaller chunk size of 1k saves ~23\% memory consumption with W4A16-K8V8. Similarly, a smaller KV cache provides an additional ~15\% memory savings from w4a16-k4v4. Furthermore, for real world deployment, we see the compounded benefit of adding optimized kernels, such as FlashAttention \cite{dao2022flashattention}, resulting an additional ~6\% memory savings from w4a16-k4v4-Flash. Note, it is imperative to evaluate the extent of task performance degradation under these Pareto-optimal configurations, even at greater context lengths. Given the application dependency, we leave such evaluations for 128k and beyond context lengths for future work.

\section{Conclusion}
We introduce KV Pareto, a systems-level framework for evaluating memory-accuracy tradeoffs in long-context LLMs. KV Pareto jointly considers prefill chunking, 4-bit weight quantization, and KV cache quantization across multiple precision levels, enabling practitioners to identify the pareto-optimal configurations for edge deployment scenarios, where maximum memory savings are needed for efficient inference. We specifically focus on optimization techniques that are lightweight and do not require out-of-box training for scalability to diverse LLMs. Our results highlight that pareto-optimal configurations are model dependent, and that our framework's chosen configurations work well in long context scenarios, as well as shorter context scenarios. Overall, KV Pareto finds optimal configurations with a 68-78\% total memory savings with 1-3\% long-context task accuracy loss. 

\section*{Limitations}
Our Pareto-optimal configurations currently use a fixed chunk size, focusing on the impact of 
enabling prefill chunking, varying KV cache quantization and weight quantization. At 128k context length, our results show that chunk size plays a critical role in performance, suggesting that future work should explore dynamic chunk sizing within the KV Pareto frontier search.
Additionally, future work should consider improving the robustness of KV cache quantization, beyond using RTN quantization. Specifically, future work should consider the inclusion of Hessian rotations, similar to QuaRot \cite{ashkboos2024quarot}, and SpinQuant \cite{liu2024spinquant}, to improve KV cache quantization and push the frontier of KV Pareto. Also, while we evaluate int8, int4 and int2 KV quantization (including mixed-precision variants), future work should expand to other quantization schemes that are adaptive and layer-specific \cite{qpruningkv, sqkv}. Additionally, while prefill chunking reduces peak memory consumption, it can introduce additional latency due to repeated KV cache writes, compared to a single-pass prefill. Future work should add latency as an additional optimization criteria in KV Pareto and analyze the frontiers' latency tradeoffs. Lastly, future work should consider the generalizability and applicability of KV Pareto to mixed model architectures such as Granite \cite{granite2024granite} or LFM2\footnote{https://www.liquid.ai/models}.


\bibliography{references}

\clearpage
\appendix
\section*{Appendix}

\section{KV Cache Growth}
\label{sec:background-kv-cache-growth}

\begin{figure}
    \centering
    \includegraphics[width=1.0\linewidth]{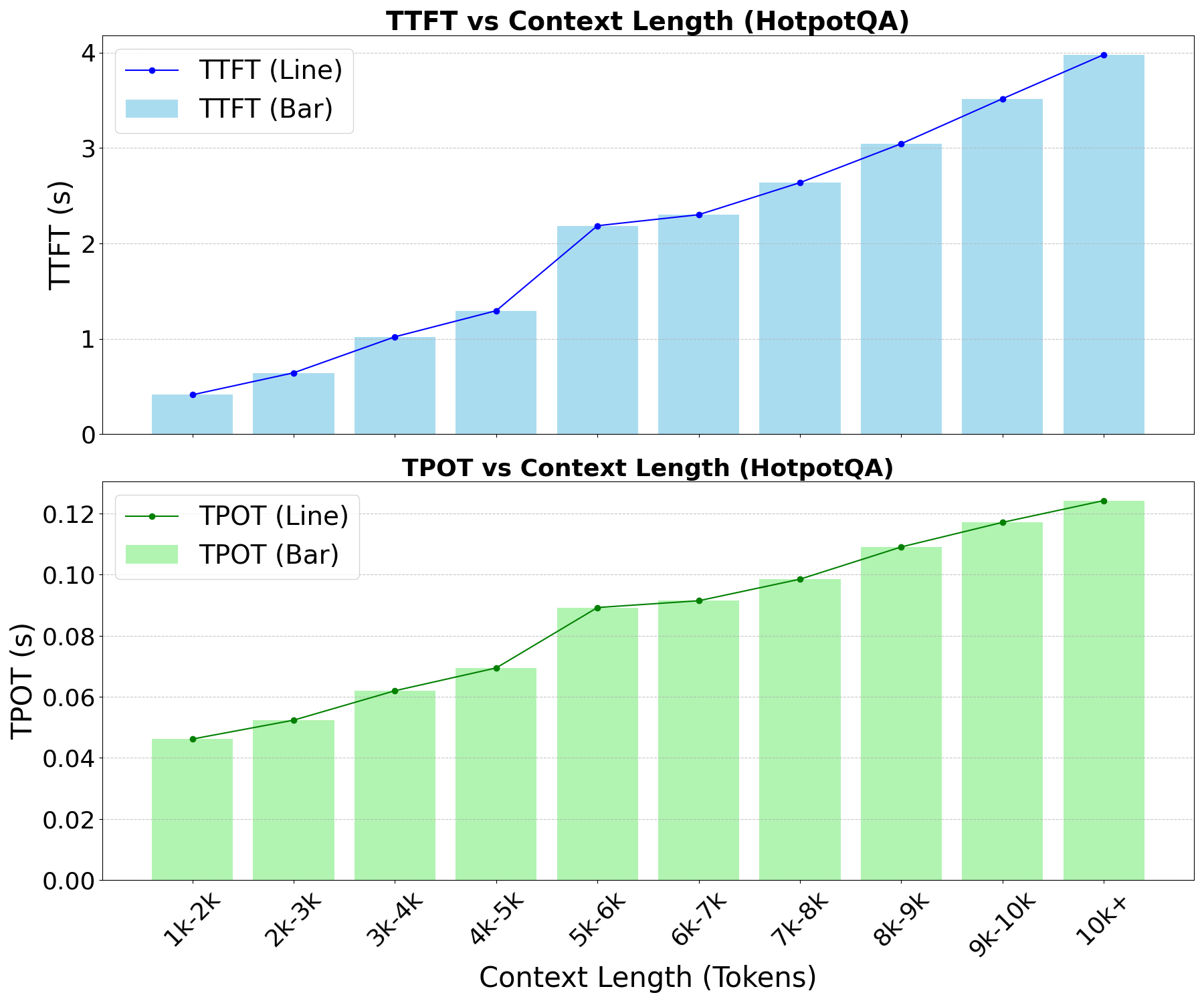}
    \caption{TPOT and TTFT curves on the the HotpotQA dataset, showcasing the bottleneck of a growing KV cache at longer contexts.}
    \label{fig:ttft_tpot}
\end{figure}

Figure \ref{fig:ttft_tpot} show how KV cache growth increases TTFT (time to first token) and TPOT (time per output token) as the context length increases, on a long context task (HotpotQA) \cite{yang2018hotpotqadatasetdiverseexplainable}, ultimately increasing inference latency. These increases arise because the prefill phase in LLMs is compute bound and incurs peak memory usage due to KV cache initialization, while the decode phase is memory-bound due to repeated KV cache access \cite{patwari2025forecasting}. 

\section{Prefill Chunking Ablations}
To study the effect of variation in chunksize on task performance, we evaluated the long context performance on chunksizes ranging from {64, 128, 256, 512, 1024}. Overall, we notice that variation in chunksize shows no impact on performance. In this table, we show ablations using the $w16a16\_k16v16$ configuration. From these results, we select a chunksize of 256 for all further experiments. 
\label{sec:PC-ablations}
\begin{table}[ht]
\centering
\small 
\begin{tabular}{@{} c cc @{}}
\toprule
\multicolumn{3}{c}{\textbf{longbench Mistral v0.2 instruct}} \\
\midrule
\textbf{prefill} & \textbf{hotpotqa} & \textbf{qasper} \\
\midrule
64   & 36.62 & 29.68 \\
128  & 37.10 & 29.30 \\
256  & 36.62 & 29.42 \\
512  & 36.76 & 29.51 \\
1024 & 36.61 & 29.21 \\
\bottomrule
\end{tabular}
\caption{HotpotQA and Qasper scores for different chunksizes for chunked prefill. Variation in chunksize does not affect task accuracy.}
\end{table}

\section{RTN Quantization Details}
\label{sec:RTN-eqs}
Round-To-Nearest Quantization (RTN) can be defined with the following.  
Let the K and V tensors have shape: \((B, H, N, D) \)
where \( B \) is batch size, \( H \) is number of attention heads, \( N \) is sequence length, and \( D \) is head dimension. A quantized tensor $q_{T}$ via RTN quantization can be defined as:
\begin{equation}
    q_{T} = round\left\lfloor \frac{T}{s} \right\rceil + z
\end{equation}
where, scale $s$ and zero point $z$ are defined follows, where $q_{min}$ and $q_{max}$ are the integer range of the target quantization:
\begin{equation}
    s = \frac{\max(T) - \min(T)}{q_{\text{max}} - q_{\text{min}}}
\end{equation}
and 
\begin{equation}
    z = round\left\lfloor q_{\text{min}} -\frac{\min(T_q)}{s}\right\rceil
\end{equation}
Similarly, for the QDQ process, de-quantization is performed as follows:
\begin{equation}
    T \approx \left\lfloor {q_T -z}\right\rceil * s
\end{equation}

\begin{figure}[t!]
    \centering
    \includegraphics[width=1.0\linewidth]{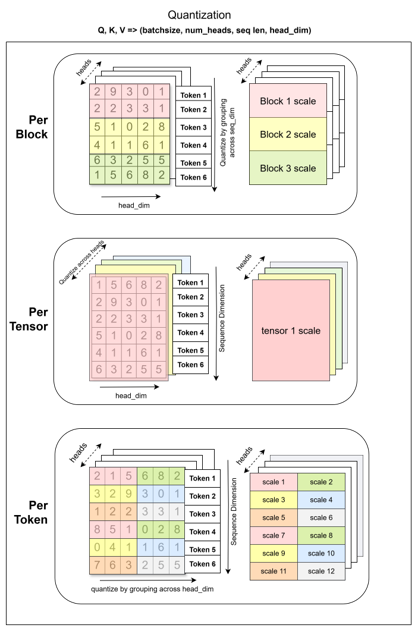}
    \caption{Illustration of KV quantization granularities.}
    \label{fig:kvquant_granularities}
\end{figure}

\paragraph{Per-token group-wise} Each token's representation is quantized independently across the heads. For each token \( t \in [1, N] \), the head dimension $D$ is divided into $G$ groups of equal size and $T \in {R}^{B \times H \times N \times \frac{D}{G} \times G}$. Scales and zero points are calculated for each group $g \in G$.
\paragraph{Per-sequence group wise} Tokens within the same sequence group share quantization parameters. Specifically, the entire sequence dimension $N$ is broken into $G$ groups of equal size and scales and zero points are calculated for each group $g \in G$.
\paragraph{Per-tensor} This represents the coarsest granularity where the entire tensor is globally quantized. Specifically, a single scale and zero point is calculated for the entire tensor $T \in {R}^{B \times H \times N \times D}$.

Figure \ref{fig:kvquant_granularities} provides a diagramatic explanation of the aforementioned granularities.

\section{KV Cache Quantization Ablations}
\label{sec:kv-quant-ablations}
We evaluated multiple KV cache quantization granularities, as outlined in Table \ref{tab:granularity_kv_longbench}, including per-token, per-block, and per tensor quantization, considering int4 and int8 precision, to isolate the effect of granularity on task accuracy. For per-token and per-block quantizations, we further ablated group sizes in the range of $\{32, 64, 128\}$. Table \ref{tab:granularity_kv_longbench}, shows that per-token quantization yields the best performance compared to per-tensor and per-block. Additionally, a group size of 32 yields the best task performance for Qwen 3B, while a group size of 64 yields the best performance for Mistral 7B. Using this information, we leverage per-token quantization with group size 32 for the the smaller models in KV Pareto (Qwen 3B and Llama3.2 3B), and per token quantization with group size 64  for the larger models (Mistral 7B and Llama3.2 8B).


{
\setlength{\tabcolsep}{1pt}
\begin{table}[t]
\centering
\scriptsize
\begin{tabularx}{\columnwidth}{l c c c *{2}{X} c *{2}{X}}
\toprule
\textbf{Granularity} & \textbf{blocksize} & \textbf{KV bits} & \multicolumn{2}{c}{\textbf{LongBench}} &
\textbf{KV bits} & \multicolumn{2}{c}{\textbf{LongBench}} \\
\hline
& \textbf{} & \textbf{} & \textbf{hotpotqa} & \textbf{qasper} &
\textbf{} & \textbf{hotpotqa} & \textbf{qasper} \\
\midrule
\multicolumn{8}{l}{\textbf{Mistral 7b}} \\
-  & -   & bf16 & 50.28  & 39.73   & bf16 & 50.28  & 39.73  \\
\midrule
Per tensor           & -   & int8 & 44.57  & 36.52  & int4 & 40.05 & 29.50  \\
Per block  & 32  & int8 & 44.81 & 40.74 & int4 & 47.36 & 38.01 \\
Per block  & 64  & int8 & 45.21 & 38.87 & int4 & 54.01 & 31.95 \\
Per block  & 128 & int8 & 47.21 & 39.75 & int4 & 48.03 & 32.47 \\
Per token  & 32  & int8 & 47.21 & 38.94 & int4 & 50.42 & 36.21 \\
Per token  & \textbf{64}  & int8 & \textbf{46.57} & \textbf{39.53} & int4 & \textbf{48.21} & \textbf{37.03} \\
Per token  & 128  & int8 & 46.57 & 39.53 & int4 & 48.21 & 37.03 \\

\midrule
\multicolumn{8}{l}{\textbf{Qwen 3b}} \\
-  & -   & bf16 & 44.59  & 37.38  & bf16 & 44.59  & 37.38   \\
\midrule
Per tensor & -   & int8 & 42.19 & 34.69  & int4 & 33.96 & 13.92 \\
Per block  & 32  & int8 & 45.68 & 35.58 & int4 & 42.45 & 25.18 \\
Per block  & 64  & int8 & 41.99 & 36.36 & int4 & 39.87 & 26.96 \\
Per block  & 128 & int8 & 43.49 & 35.88 & int4 & 31.19 & 25.01 \\
Per token  & \textbf{32}  & \textbf{int8} & \textbf{43.63} & \textbf{36.75} & int4 & \textbf{45.37} & \textbf{34.91} \\
Per token  & 64  & int8 & 41.63 & 36.24 & int4 & 37.96 & 27.56 \\
Per token  & 128 & int8 & 41.39 & 36.01 & int4 & 40.01 & 33.26 \\
\bottomrule
\end{tabularx}

\caption{Granularity-wise KV precision and LongBench scores for Mistral 7b and Qwen 3b.}
\label{tab:granularity_kv_longbench}
\end{table}
}

\section{K-smoothing Method}
K-smoothing is inspired from SageAttention \cite{zhang2025sageattention} where mean-smoothing is applied to the $K$ tensor. Specifically, we apply the following:
\label{sec:k-smoothing-method}
\begin{equation}
\label{eq:k_smooth}
    \tilde{K}_{b,i,d} = K_{b,i,d} - \frac{1}{L} \sum_{j=1}^{L} Kk_{b,j,d}
\end{equation}
where  $K$ \( \in \mathbb{}b{R}^{B \times L \times D} \)  is the original tensor, \(\tilde{K}\) is the mean-centered tensor, $B$ is the batch size, $L$  is the sequence length (dimension over which mean is computed), $D$ is the feature or head dimension, $b$ indexes the batch, $i$ indexes the sequence position, $d$ indexes the feature dimension.

\subsection{K-smoothing Ablations}
\label{sec:ksmoothing}

Our ablation studies show that subtracting $K_{mean}$ (averaging along sequence dimension) from $K$ for per-token quantization gives the best configuration for smoothing. Overall, this shows the necessity of K smoothing for lower precision (int4) support. 

\begin{table}[ht]
\centering
\small 
\begin{tabular}{@{} c c c @{}}
\toprule
\multicolumn{3}{c}{\textbf{K smoothing: qwen3b}} \\
\midrule
precision & averaging across & hotpotqa \\
\midrule
int8 & No smoothing  & 44.77 \\
int8 & $head\_dim$  & 44.46 \\
int8 & $seq\_len$ & \textbf{46.15} \\
\midrule
int4 & No smoothing  & gibberish \\
int4 & $head\_dim$  & gibberish \\
int4 & $seq\_len$ & \textbf{41.69} \\
\bottomrule
\end{tabular}
\caption{Results for K smoothing by subtracting mean across various dimensions, for int4 and int8. Including K smoothing improves results significantly.}
\end{table}

\section{Evaluation Dataset Details}
\label{sec:eval-datasets}
We evaluate KV Pareto on long context datasets from LongBench \cite{bai2024longbench}, specifically HotpotQA \cite{yang2018hotpotqadatasetdiverseexplainable} and Qasper \cite{qasper}. Both Qasper and HotpotQA  
evaluate multi-document QA and single-document QA using F1 scores. The average prompt length in HotpotQA is 9k, whereas the average prompt length in Qasper is 4k. We additionally evaluate on the Needle-in-a-haystack (NIAH) \cite{NIAH} which evaluates text retrieval (needle), in large document scenarios. The NIAH benchmark supports up to 32k context length. 

We also evaluate on GSM8k \cite{gsm8k} and MMLU \cite{mmlu} tasks which are not considered long context tasks to ensure minimal task performance degradation on these standard evaluation tasks. For both GSM8k \cite{gsm8k} and MMLU \cite{mmlu} evaluations, we leveraged LM-Eval-Harness \cite{eval-harness}, and specifically set the evaluation sample size to 50 across all subjects for MMLU. 

\section{Longbench Pareto curves}
\label{sec:add-pareto-curves}

\begin{figure}
    \includegraphics[width=1.0\linewidth]{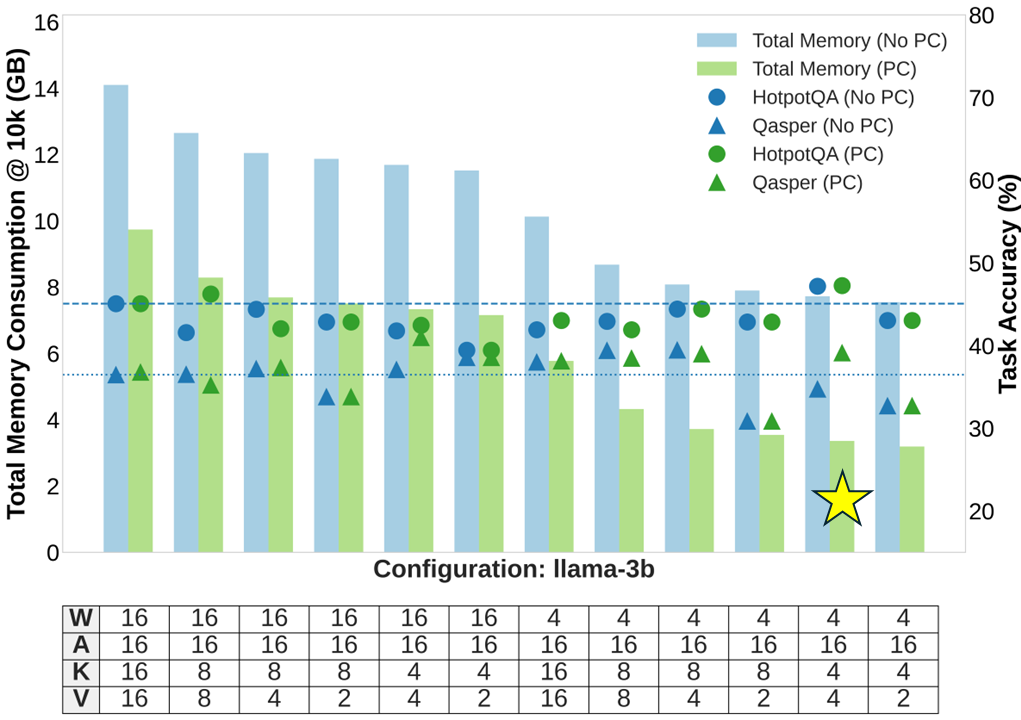}
    \caption{Llama-3.2-3B-Instruct's pareto optimal search.}
    \label{fig:pareto_slide4}
\end{figure}

Figure \ref{fig:pareto_slide4} shows an additional pareto search from our KV Pareto framework for the llama-3.2-3b-instruct model. T pareto-optimal solution is at W4A16\_K4V4 configuration. 

\section{NIAH results}
\label{sec:niah-results}
\begin{figure}[t!]
    \centering
    \captionsetup{justification=centering}

    \begin{subfigure}[b]{0.49\textwidth}
        \includegraphics[width=\linewidth]{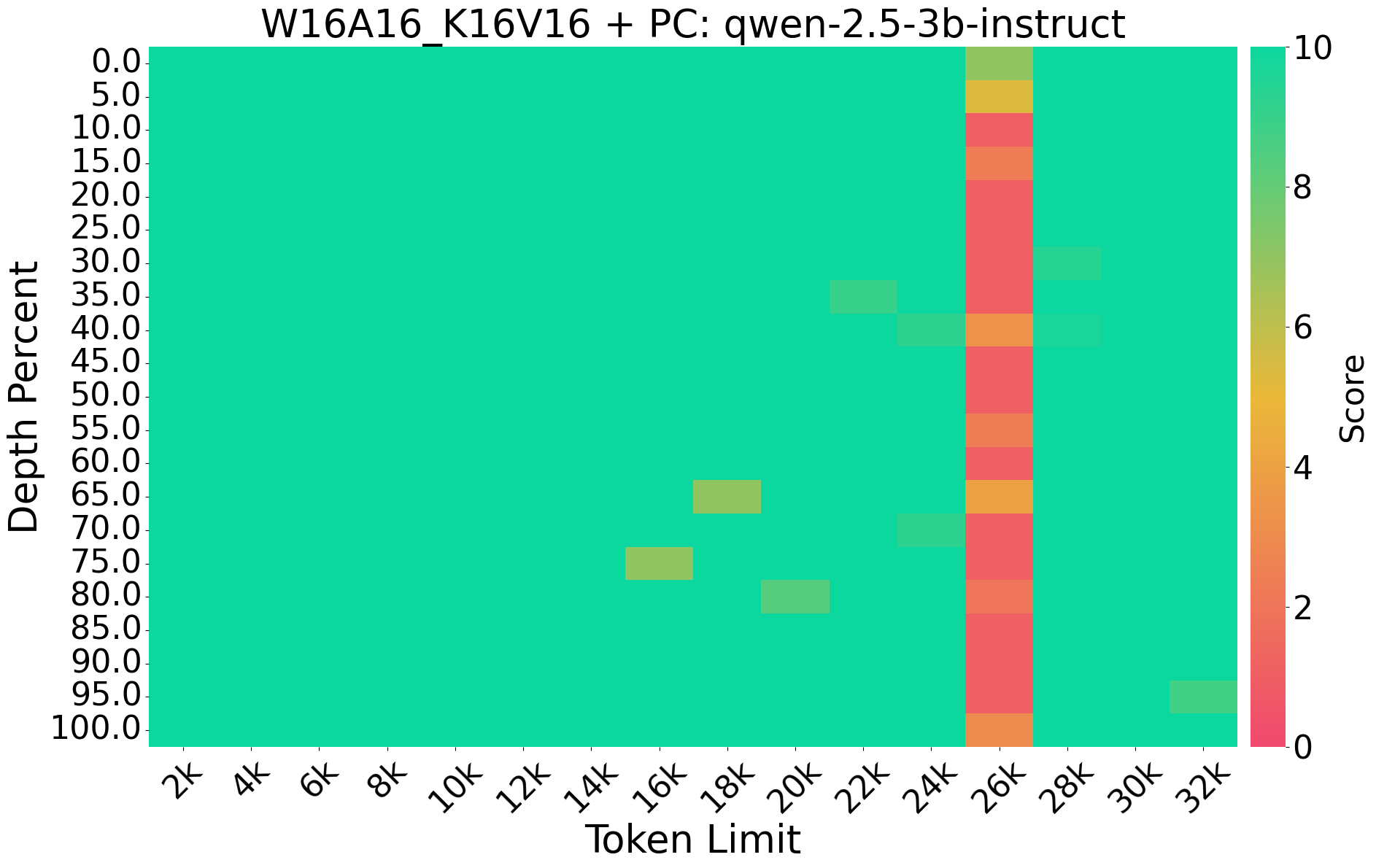}
        \caption{W16A16\_K16V16 retrieval scores for Qwen-2.5-3B-I}
        \label{fig:niah3}
    \end{subfigure}
    \hfill
    \begin{subfigure}[b]{0.49\textwidth}
        \includegraphics[width=\linewidth]{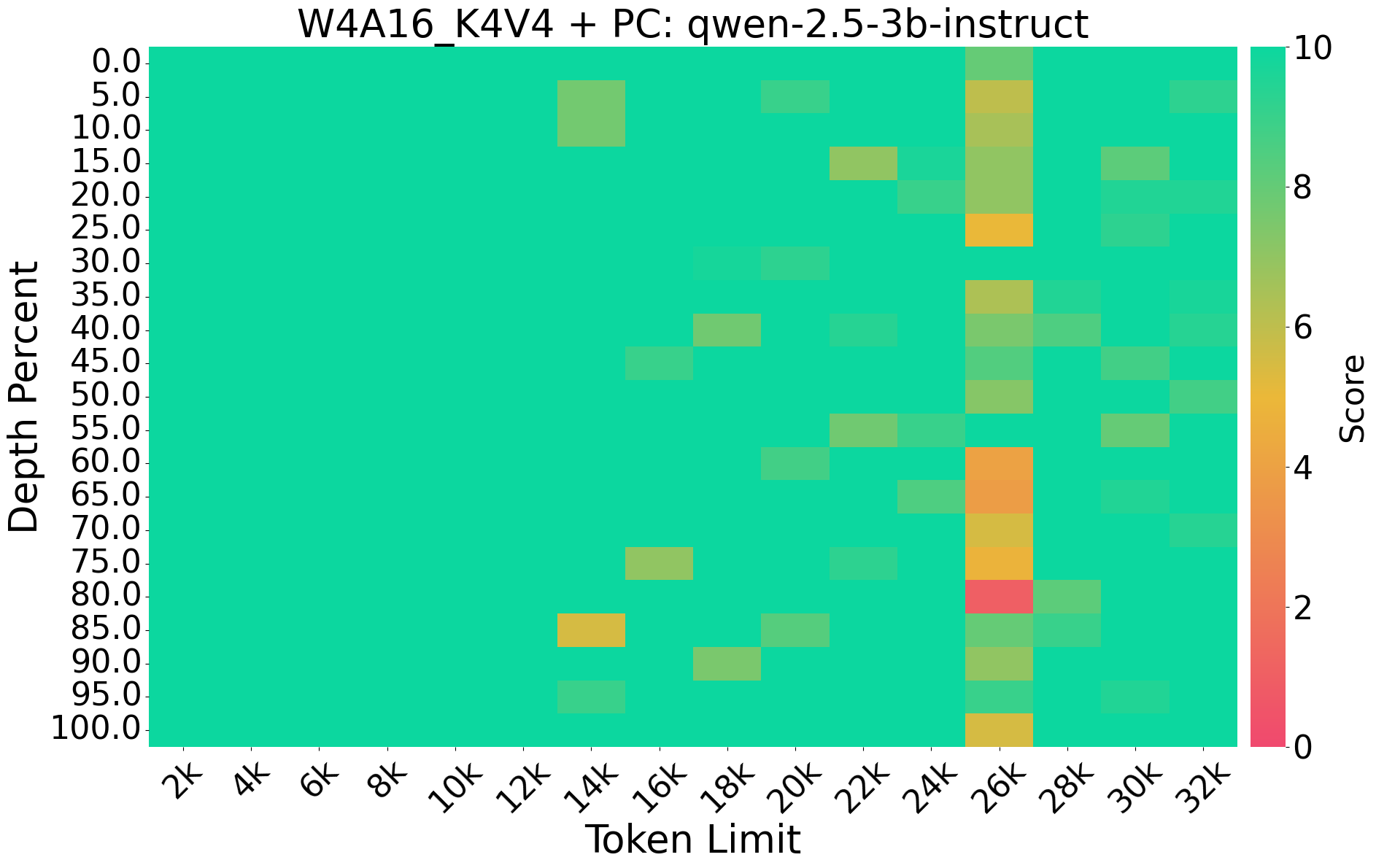}
        \caption{W4A16\_K4V4 retrieval scores for Qwen-2.5-3B-I} 
        \label{fig:niah4}
    \end{subfigure}

    \caption{NIAH performance on selected configurations.}
    \label{fig:niah_grid2}
\end{figure}

We also assess information retrieval performance using the Needle in a Haystack benchmark on the Qwen-2.5-3B-Instruct model. Figure \ref{fig:niah_grid2} illustrates retrieval accuracy up to a 32k context length. Notably, the baseline results show poor performance at 26k tokens, which may be due to model-specific behavior. The $w4a16\_k4v4$ configuration maintains acceptable performance up to 14k context length.

\section{Memory Consumption Approximation Details}
\label{sec:mem-util-calc}
The memory consumed by the model is a sum of model parameters, KV cache size and peak activation memory. Model parameters are calculated by counting the parameters and corresponding datatypes. KV cache is calculated as described in Section \ref{background}. The peak activation memory is dominated by either lm\_head layer or by the MHA depending on the operating conditions.
\begin{equation}
    mem_{MHA} = \begin{cases}
   n_h \times M \times M & \text{if SDPA,}\\
   n_h \times c \times M & \text{if SDPA, PC}\\
   b_q^2 + ( b_{kv}^2 \times 2 ) +\Delta & \text{if Flash-MHA}\\
\end{cases}
\end{equation}

\begin{equation}
   mem_{lm\_head} = M \times vocab\_size
\end{equation}

\begin{equation}
   mem_{peak} = max(mem_{MHA}, mem_{lm\_head})
\end{equation}

where $n_h$ denotes number of attention heads, $b_q$ and $b_{kv}$ are block sizes in Flash Attention kernel, $M$ is the total sequence length and $c$ is the chunk size.

\begin{table*}[!t]
\centering
\captionsetup{justification=centering}
\large 
\setlength{\tabcolsep}{6pt} 
\renewcommand{\arraystretch}{1.1} 
\resizebox{\textwidth}{!}{%
\begin{tabular}{@{} l c l c c @{}}
\toprule
\textbf{Model Name} & \textbf{bf16 baseline} & \textbf{Pareto} & \textbf{Memory @} & \textbf{Memory} \\
& \textbf{memory (GB)} & \textbf{Optimality} & \textbf{ Optimality (GB)} & \textbf{reduction \%} \\
\midrule
Qwen 2.5 3B Instruct      & 11.49 & W4A16\_K4V4 + PC & 3.10 & \textbf{73\%} \\
Llama 3.2 3B Instruct     & 14.10 & W4A16\_K4V4 + PC & 3.36 & \textbf{76\%} \\
Qwen 2.5 7B Instruct      & 24.90 & W4A16\_K8V8 + PC & 7.74 & \textbf{68\%} \\
Llama 3.1 8B Instruct     & 26.91 & W4A16\_K8V2 + PC & 6.83 & \textbf{75\%} \\
Mistral v0.3 7B Instruct  & 24.34 & W4A16\_K4V4 + PC & 5.52 & \textbf{78\%} \\
\bottomrule
\end{tabular}
}
\caption{Pareto-optimal memory configurations for different LLMs.}
\label{tab:pareto_memory}
\end{table*}

\end{document}